\documentclass[fleqn]{llncs}
\usepackage{epsf}
\usepackage{amsmath}
\usepackage{authblk}
\newcommand{\superscript}[1]{\ensuremath{^{\textrm{#1}}}} 
\def\wu{\superscript{1}} 
\def\wg{\superscript{2}}
\def\wp{\superscript{3}}
\def\comma{\superscript{,}}
\begin{document}

\title{End-to-End Relation Extraction using Markov Logic Networks}

\author{Sachin Pawar\wu\comma\wg}
\author{Pushpak Bhattacharya\wg\comma\wp}
\author{Girish K. Palshikar\wu}
\affil{\tt sachin7.p@tcs.com, pb@cse.iitb.ac.in, gk.palshikar@tcs.com}
\affil{\wu TCS Research, Tata Consultancy Services, Pune-411013, India}
\affil{\wg Dept. of CSE, Indian Institute of Technology Bombay, Mumbai-400076, India}
\affil{\wp Indian Institute of Technology Patna, Patna-801103, India}
\institute{}

\maketitle

\begin{abstract}
The task of end-to-end relation extraction consists of two sub-tasks: i) identifying entity mentions along with their types and ii) recognizing semantic relations among the entity mention pairs.
It has been shown that for better performance, it is necessary to address these two sub-tasks jointly~\cite{roth2004linear,li2014incremental}. We propose an approach for simultaneous extraction of entity mentions and relations in a sentence, by using inference in Markov Logic Networks (MLN)~\cite{richardson2006markov}. We learn three different classifiers : i) local entity classifier,  ii) local relation classifier and iii) ``pipeline'' relation classifier which uses predictions of the local entity classifier. Predictions of these classifiers may be inconsistent with each other. We represent these predictions along with some domain knowledge using weighted first-order logic rules in an MLN and perform joint inference over the MLN to obtain a global output with minimum inconsistencies. Experiments on the ACE (Automatic Content  Extraction) 2004 dataset demonstrate that our approach of joint extraction using MLNs outperforms the baselines of individual classifiers. Our end-to-end relation extraction performance is better than 2 out of 3 previous results reported on the ACE 2004 dataset.

\end{abstract}

\section{Introduction}
Real world entities are referred in natural language sentences through {\em entity mentions} and these are often linked through meaningful {\em relations}. The task of end-to-end relation extraction consists of two sub-tasks: entity extraction and relation extraction. The sub-task of {\em entity extraction} deals with identifying entity mentions and determining their entity types. The other task of {\em relation extraction} deals with identifying whether any semantic relation exists between any two mentions in a sentence and also determining the relation type if it exists. In this paper, we refer to {\em entity extraction} and {\em relation extraction} tasks as defined by the Automatic Content Extraction (ACE) program~\cite{doddington2004automatic} under the EDT (Entity Detection and Tracking) and RDC (Relation Detection and Characterization) tasks, respectively. ACE standard defined 7 entity types~\footnote{\url{https://www.ldc.upenn.edu/sites/www.ldc.upenn.edu/files/english-edt-v4.2.6.pdf}}: PER (person), ORG (organization), LOC (location), GPE (geo-political entity), FAC (facility), VEH (vehicle) and WEA (weapon). It also defined 7 coarse level relation types~\footnote{\url{https://www.ldc.upenn.edu/sites/www.ldc.upenn.edu/files/english-rdc-v4.3.2.PDF}}: EMP-ORG (employment), PER-SOC (personal/social), PHYS (physical), GPE-AFF (GPE affiliation), OTHER-AFF (PER/ORG affiliation), ART (agent-artifact) and DISC (discourse). 

Compared to the work (refer the surveys~\cite{nadeau2007survey,palshikar2013techniques}) in Named Entity Recognition (NER), there are relatively few attempts~\cite{florian2006factorizing,florian2010improving,li2014incremental,lu2015joint} to address the more general entity extraction problem. NER extracts only named mentions (e.g. \texttt{John Smith}, \texttt{Walmart}) whereas entity extraction is expected to also identify common noun and pronoun mentions (e.g. \texttt{company}, \texttt{leader}, \texttt{it}, \texttt{they}) and their entity types. This task is more challenging than NER because entity type of mentions like {\tt leader} or {\tt they} may vary from sentence to sentence depending on which real life entity they are referring to in that sentence. For example, entity type of {\tt leader} would be PER in the sentence {\tt John Smith was elected as the leader of the Socialist Party} whereas its entity type would be ORG in the sentence {\tt Pepsi is a market leader in its segment.}

There has been a lot of work for relation extraction like Zhou et al.~\cite{zhou2005}, Jiang and Zhai~\cite{jiang2007}, Bunescu and Mooney~\cite{bunescu2005} and Qian et al.~\cite{qian2008exploiting}. All of these approaches assume that the boundaries and the types of entity mentions are already known. Several features based on this information are used for relation prediction. In order to use such relation extraction systems, there should be separate entity extraction system whose output acts as an input for relation extraction. In such a ``pipeline'' method, the errors are propagated from first phase (entity extraction) to second phase (relation extraction) affecting the overall relation extraction performance. Another major disadvantage of the ``pipeline'' method is that it facilitates only one-way {\em information flow}, i.e. the knowledge about entities is used for relation extraction but not vice versa. However, the knowledge about relations can help in correcting some entity extraction errors. 

In order to overcome these problems, we propose an approach which uses inference in Markov Logic Networks (MLN) for simultaneous extraction of entities and relations in a sentence. This approach facilitates two-way {\em information flow}. MLNs combine first-order logic and probabilistic graphical models in a single representation. An MLN contains a set of first-order logic rules, and each rule is associated with a weight. The fewer rules a world violates, the more probable it is. Also, higher the weight of a rule, greater is the probability of a world that satisfies the rule compared to the one that does not. In our approach, three separate classifiers are learned: a local entity classifier, a local relation classifier and a ``pipeline'' relation classifier which uses predictions of the local entity classifier. Predictions of these classifiers along with other domain knowledge are represented using weighted first-order logic rules in an MLN. Joint inference over this MLN is then performed to get a final output with least possible contradictions or inconsistencies among the individual classifiers.



The specific contributions of this work are : i) a novel approach for joint extraction of entity mentions and relations using inference in MLNs and ii) easy and compact representation of the domain knowledge using first-order logic rules in MLNs. The rest of the paper is organized as follows. Section 2 describes some background and necessary building blocks for our approach. Section 3 describes our approach in detail and Section 4 describes the working of our approach through an example. Experimental results are presented in Section 5. Related work is then described briefly in Section 6. Finally we conclude in Section 7 with brief discussion about the future work.

\section{Building Blocks for Our Approach}

\subsection{Markov Logic Networks}
Markov Logic Networks (MLN) which were proposed by Richardson and Domingos~\cite{richardson2006markov}, combine first-order logic and probabilistic graphical models in a single representation. Formally, a Markov Logic Network $L$ is defined as a set of pairs $(F_i,w_i)$, where each $F_i$ is a formula in first-order logic with a real weight $w_i$. Along with a finite set of constants $C=\{C_1, C_2,\cdots, C_{|C|}\}$, it defines a Markov Network $M_{L,C}$ as follows:
\begin{enumerate}
\item $M_{L,C}$ contains one binary node for each possible grounding of each predicate appearing in $L$. The value of the node is $1$ if the ground atom is true, and $0$ otherwise.
\item $M_{L,C}$ contains one feature for each possible grounding of each formula $F_i$ in $L$. The value of this feature is $1$ if the ground formula is true, and $0$ otherwise. The weight of the feature is the $w_i$ associated with $F_i$ in $L$.
\end{enumerate}
The probability distribution of random variable $X$ over possible worlds $x$ specified by Markov Network $M_{L,C}$ is given by,

\vspace{-4mm}
{\footnotesize
\begin{equation}
P(X=x) = \frac{1}{Z} \exp \left(\sum_i w_i n_i(x)\right)
\label{eq:mln_prob}
\end{equation}
}
where $n_i(x)$ is the number of true groundings of $F_i$ in $x$ and $Z$ is the partition function. 
MLN can be used to find probability of a formula (say $F_1$) being true, given some other formula (say $F_2$) is true. 

\vspace{-4mm}
{\footnotesize
\begin{equation*}
P(F_1|F_2,M_{L,C}) = \frac{P(F_1 \wedge F_2|M_{L,C})}{P(F_2|M_{L,C})} = \frac{\sum_{x\in X_{F_1} \cap X_{F_2}} P(X=x|M_{L,C})}{\sum_{x\in X_{F_1}} P(X=x|M_{L,C})}
\end{equation*}
}
where $X_{F_i}$ represents the set of worlds where $F_i$ holds and $P(X=x|M_{L,C})$ is computed using the equation~\ref{eq:mln_prob}. 

\subsection{Identifying Entity Mention Candidates}
It is necessary to first identify the span (or boundaries) of each entity mention~\footnote{We consider the ``head'' extent of a mention defined by ACE standard as the entity mention so that all the valid entity mentions are always non-overlapping.} in a given sentence. We model this as a sequence labelling problem.
A sentence is a sequence of words and each word in a sentence is assigned a label indicating whether that word belongs to any entity mention or not. We use BIO encoding for this purpose.
\begin{itemize}
\item \textbf{O:} Label for the words which are not part of any entity mention
\item \textbf{B:} Label for the first word of entity mentions
\item \textbf{I:} Label for the subsequent words (except the first word) of entity mentions
\end{itemize}
We employ the Conditional Random Field (CRF) model~\cite{lafferty2001conditional}, which is trained in a supervised manner.
Given any new sentence, we use the trained CRF model to predict the $2$ most probable label sequences as follows:
\begin{center}\footnotesize
$S_1$:{\tt A}/O {\tt Palestinian}/B {\tt Council}/B {\tt member}/B {\tt says}/O {\tt anger}/O {\tt is}/O {\tt growing}/O {\tt .}/O\\
$S_2$:{\tt A}/O {\tt Palestinian}/B {\tt Council}/I {\tt member}/B {\tt says}/O {\tt anger}/O {\tt is}/O {\tt growing}/O {\tt .}/O
\end{center}
In this sentence, entity mention candidates from the topmost sequence are {\tt Palestinian}, {\tt Council} and {\tt member}. Entity mention candidate {\tt Palestinian Council} is generated from the second sequence. Generally, the candidates generated from the most probable sequence are more likely to be valid entity mentions. The candidates generated from the second most probable sequence are considered valid entity mentions only if they satisfy certain constraints. These constraints are applied in the form of first-order logic rules in MLNs and will be explained later. A special entity type $NONE$ is assigned to a candidate entity mention if it is an invalid entity mention.
\subsection{Local Entity Classifier}
The local entity classifier is used to predict the most probable entity type for each candidate entity mention in a given sentence. This classifier is referred to as ``local'' as it takes an independent decision for each entity mention irrespective of its relation with other mentions. 
A Maximum Entropy Classifier is trained in a supervised manner which captures the characteristics of each entity mention $E$ using following features:\\
1. \textbf{Lexical Features:} Head word and other words in $E$, words preceding and succeeding $E$ in the sentence\\
2. \textbf{Syntactic Features:} POS tags of the head word and other words in $E$, POS tags of the words preceding and succeeding $E$, parent of head word of $E$ in the dependency tree and also the dependency relation with the parent\\
3. \textbf{Semantic Features:} WordNet category (if any) of the head word of $E$. Some specific synsets in the WordNet (e.g. \texttt{person}, \texttt{location}, \texttt{vehicle}) are marked as possible ``categories'' and if any word is direct or indirect hypernym of such synsets, it is said to be falling in the corresponding ``category''.\\
As this classifier is trained using only the valid entity mentions in the training data, it always predicts one of the 7 ACE entity types and never predicts the $NONE$ type. 

\subsection{Local Relation Classifier}
The local relation classifier is used to predict the most probable relation type for each pair of candidate entity mentions in a given sentence. This classifier is referred to as ``local'' as it takes an independent decision for each pair of entity mentions irrespective of their entity types. 

In addition to ACE 2004 relation types, it considers two special relation types ``NULL'' (indicating that no semantic relation holds) and ``IDN'' (representing intra-sentence co-references). 
In the sentence \texttt{Pepsi is a market leader}, the entity mentions {\tt leader} and {\tt Pepsi} are co-references and hence we add the IDN relation between these mentions. With the help of IDN (identity) relation type, information about intra-sentence co-references can be incorporated in a principled way without using an external co-reference resolution system. Also, more number of entity mentions get involved in at least one relation, resulting in better entity extraction performance. For example, in the ACE 2004 dataset, there are $22718$ entity mentions and $4328$ relation instances resulting in only $7604$ entity mentions involved in at least one relation. Considering the IDN relation, number of relation instances increases to $12060$ covering $14930$ entity mentions.

A Maximum Entropy Classifier is used which captures the characteristics of each entity mention pair $(E_1,E_2)$ with the help of following features:\\
1. \textbf{Lexical Features:} Head words and other words of $E_1$ \& $E_2$, words preceding and succeeding $E_1$ \& $E_2$ in the sentence\\
2. \textbf{Syntactic Features:} POS tags of the head word and other words in $E_1$ \& $E_2$, POS tags of the words preceding and succeeding $E_1$ \& $E_2$, parents of head words of $E_1$ \& $E_2$ in the dependency tree and also the dependency relations with the parents, path connecting $E_1$ \& $E_2$ in the dependency tree, their common ancestor in the dependency tree\\
3. \textbf{Semantic Features:} WordNet categories (if any) of the head words of $E_1$ \& $E_2$, the common ancestor and other words on the path connecting $E_1$ \& $E_2$ in the dependency tree of the sentence, syntactico-semantic structures identified in Chan and Roth~\cite{chan2011exploiting}.

\subsection{Pipeline Relation Classifier}
Unlike the local relation classifier, the ``pipeline'' relation classifier is dependent on the output of the local entity classifier.
It uses following features in addition to the features used by the local relation classifier.\\
1. Entity types of $E_1$ and $E_2$ as predicted by the local entity classifier\\
2. Concatenation of entity types of $E_1$ and $E_2$\\
3. A binary feature indicating whether the types of $E_1$ and $E_2$ are same or not.

This classifier is referred to as a ``pipeline'' classifier because of unidirectional {\em information flow}. In other words, the knowledge about types of entity mentions is used by the relation classifier but not vice versa.

\section{Joint Extraction using Inference in MLNs}
\subsection{Motivation}
As described in the previous section, we have 3 classifiers producing various predictions about entity types and relation types. These decisions may be inconsistent, i.e. relation type predicted by the local relation classifier may not be compatible with the entity types predicted by the local entity classifiers. Also, there may be contradiction in predictions of local relation classifier and ``pipeline'' relation classifier. Our aim is to take predictions of these classifiers as input and make a global prediction which minimizes such inconsistencies. MLN provides a perfect framework for this, where we can represent predictions of individual classifiers as first-order logic rules where weights of these rules are proportional to the prediction probabilities (soft constraints). Also, the consistency constraints among the relation types and entity types can be represented in the form of first-order logic rules with infinite weights (hard constraints). Now, the inference in such an MLN will provide a globally consistent output with maximum weighted satisfiability of the rules. The detailed explanation is provided in subsequent sections about how the first-order logic rules are created and how the corresponding weights are set.
\subsection{Domains and Predicates}
We specify one MLN for a sentence, i.e. for all candidate entity mentions and possible relation instances in a sentence. 
The software package used for inference in MLN is Alchemy~\footnote{\url{http://alchemy.cs.washington.edu/}}.
We define 3 domains : $entity$, $etype$ and $rtype$. The $entity$ domain represents entity mentions where an unique ID is assigned to each entity mention. 
It is specified as follows in Alchemy for a sentence with $n$ entity mentions having IDs from $1$ to $n$:

\vspace{-5mm}
{\footnotesize
\begin{equation*}
entity = \{1,2,\cdots,n\}
\end{equation*}
}
The next domain $etype$ represents the set of all possible entity types and another domain $rtype$ represents the set of all possible relation types. These domains are specified in Alchemy as follows:

\vspace{-5mm}
{\footnotesize
\begin{equation*}
etype = \{PER, ORG, LOC, GPE, WEA, FAC, VEH, NONE\}
\end{equation*}
\begin{equation*}
rtype = \{EMPORG, GPEAFF, OTHERAFF, PERSOC, PHYS, ART, NULL, IDN\}
\end{equation*}
}

We define following predicates which are used for writing various first-order logic rules. The arguments for these predicates come from the above domains.\\
1. $ET(entity,etype)$: $ET(i,E)$ is true only when entity type of the entity mention $i$ is equal to $E$. It is true for one and only one entity type. It represents the entity type prediction of the local entity classifier and used as an {\em evidence} during inference.\\
2. $RTP(entity,entity,rtype)$: $RTP(i,j,R)$ is true only when type of relation between entity mentions $i$ and $j$ is equal to $R$. It is true for one and only one relation type. It represents relation type prediction of ``pipeline'' relation classifier. It is also used as an {\em evidence}.\\
3. $RTL(entity,entity,rtype):$ Similar to $RTP$ but represents relation type prediction of local relation classifier.\\
4. $ETFinal(entity,etype):$ Similar to $ET$ but represents global entity type prediction and is used as a {\em query} predicate during inference.\\
5. $RTFinal(entity,entity,rtype):$ Similar to $RTP$ but represents global relation type prediction and is used as a {\em query} predicate.

During the inference in MLN, the probabilities of all possible groundings of {\em query} predicates are computed, conditioned on the specific groundings of the {\em evidence} predicates.

\subsection{Generic Rules}
Although one MLN is created for each sentence, some first-order rules are common and they are added to MLNs of all the sentences. We refer to these rules as {\em Generic Rules}. These rules represent some universal truths about the domain and hence the weight associated with each of these rules is set to {\em infinity}. In other words, any world that violates any of these rules, is practically impossible. These rules provide an easy and effective way of incorporating the domain knowledge about entity types and relation types. For each valid combination of relation type and entity type of one of its argument, we write rules to constrain the possible entity types for the other argument. Such rules can be easily devised by going through the ACE 2004 labelling guidelines. Following are some representative examples~\footnote{All the rules can't be listed because of the space constraints.}. Note that the variables $x,y$ are universally quantified at the outermost level.\\
1. If there is an $EMPORG$ relation between two entity mentions and entity type of any mention is $PER$, then entity type of other mention can only be one of : $ORG$ or $GPE$. 

\vspace{-5mm}
{\small
\begin{eqnarray*}
RTFinal(x,y,EMPORG) \wedge ETFinal(x,PER) \Rightarrow (ETFinal(y,ORG) \vee ETFinal(y,GPE)).\\
RTFinal(x,y,EMPORG) \wedge ETFinal(y,PER) \Rightarrow (ETFinal(x,ORG) \vee ETFinal(x,GPE)).
\end{eqnarray*}
}
2. For the ``identity'' relation type $IDN$, the constraint is that the entity types of both the mentions should be same.

\vspace{-5mm}
{\footnotesize
\begin{eqnarray*}
RTFinal(x,y,IDN) \wedge ETFinal(x,z) \Rightarrow ETFinal(y,z).\\
RTFinal(x,y,IDN) \wedge ETFinal(y,z) \Rightarrow ETFinal(x,z).
\end{eqnarray*}
}
\subsection{Sentence-specific Rules}
These rules are specific to each sentence and represent the predictions by the individual baseline classifiers. Unlike the {\em Generic Rules}, these rules are added with finite weights. 

\noindent\textbf{Weight Assignment Strategies:} In order to learn the weights of various first-order logic rules, historical examples of predictions of 3 base classifiers along with gold-standard predictions would be required. Instead we chose to compute these weights by using some functions of the corresponding prediction probabilities. The work by Jain~\cite{jain2011knowledge} discussed various ways of weight assignments to represent knowledge in MLNs. In another work, Heckmann et al.~\cite{heckmann2013citation} adjusted the rule weights experimentally for citation segmentation using MLNs. On the similar lines, following two strategies are adopted for weight assignments.
\begin{enumerate}
\item \textbf{Log of Odds Ratio (LOR):} Richardson and Domingos~\cite{richardson2006markov} states that the weight of a formula $F$ is log odds between a world where $F$ is true and a world where $F$ is false. For a prediction with probability $p$, we set the weight of corresponding formula as $\log\left(\frac{p}{1-p}\right)$. Here, the penalty for violating any formula will increase logarithmically with its probability.
\item \textbf{Constant Multiplier (CM):} As per this strategy, for a prediction with probability $p$, we set the weight of corresponding formula as $K\cdot p$. Here, the penalty for violating any formula will increase linearly with its probability. We have used $K=10$ in all our experiments.
\end{enumerate}

\noindent\textbf{Rules induced by the Local Entity Classifier:} For each candidate entity mention, the entity type predicted by the local entity classifier acts as an {\em evidence} for the MLN inference. The classifier also assigns some probability to each possible entity type. For each entity mention id $i$, for each possible entity type $E$, 
 following rule is added with the weight proportional to the probability of prediction.

\vspace{-5mm}
{\footnotesize
\begin{equation*}
ET(i,E_{max}) \Leftrightarrow ETFinal(i,E)
\end{equation*}
}
Here, $E_{max}$ is the entity type predicted by the local entity classifier. 
The weights assigned to this rule as per above strategies would be $\log\left(\frac{P_e(E|i)}{1-P_e(E|i)}\right)$ and $K\cdot P_e(E|i)$, where $P_e(E|i)$ is the probability assigned to entity type $E$ for the entity mention id $i$ by the local entity classifier. 

\noindent\textbf{Rules induced by the Pipeline Relation Classifier:} For each pair of entity mentions, the relation type predicted by the ``pipeline'' classifier acts as an {\em evidence} for the MLN inference. 
For each pair of candidate entity mentions $(i,j)$, for each possible relation type $R$, 
following rule is added with the weight proportional to the probability of prediction.

\vspace{-5mm}
{\footnotesize
\begin{equation*}
RTP(i,j,R_{max}) \Leftrightarrow RTFinal(i,j,R)
\end{equation*}
}
Here, $R_{max}$ is the relation type predicted by the ``pipeline'' relation classifier.
The weights assigned to this rule would be $\log\left(\frac{wt_p\cdot P^P_r(R|i,j)}{1-P^P_r(R|i,j)}\right)$ and $K\cdot P^P_r(R|i,j)\cdot wt_p$, where $P^P_r(R|i,j)$ is the probability assigned to the relation type $R$ for the pair $(i,j)$ by the ``pipeline'' relation classifier. And $wt_p$ is the reliability of prediction of ``pipeline'' classifier, which 
indicates how confident the local entity classifier was in predicting entity types for entity mentions $i$ and $j$. We set $wt_p = P_e(E^i_{max}|i)\cdot P_e(E^j_{max}|j)$. 


\noindent\textbf{Rules induced by the Local Relation Classifier:} For each pair of candidate entity mentions, the relation type predicted by the local classifier acts as an {\em evidence} for the MLN inference. 
For each pair entity mentions $(i,j)$, for each possible relation type $R$, 
following rule is added with the weight proportional to the probability of prediction.

\vspace{-5mm}
{\footnotesize
\begin{equation*}
RTL(i,j,R^L_{max}) \Leftrightarrow RTFinal(i,j,R)
\end{equation*}
}
Here, $R^L_{max}$ is the relation type predicted by the local relation classifier.
The weights assigned to this rule would be $\log\left(\frac{P^L_r(R|i,j)}{1-P^L_r(R|i,j)}\right)$ and $K\cdot P^L_r(R|i,j)$, where $P^L_r(R|i,j)$ is the probability assigned to the relation type $R$ for the pair $(i,j)$ by the local relation classifier. 

\noindent\textbf{Rules for identifying valid/invalid entity mentions:} We generate candidate entity mentions using top $2$ most probable BIO sequences. In general, we have a high confidence that candidate mentions from the topmost sequence are valid and have a lower confidence for candidates from the second sequence. This intuition is captured by addition of following rules. For each candidate $i$ from the topmost sequence, we add $!ETFinal(i,NONE)$ with the weight $\log\left(\frac{p}{1-p}\right)$ or $K\cdot p$, based on the weighing strategy employed. Also for each candidate $i$ from the second sequence, we add $ETFinal(i,NONE)$ with the weight $\log\left(\frac{1-p}{p}\right)$ or $K\cdot(1-p)$. In both the cases, $p$ is the highest probability for any entity type predicted for that mention by the local entity classifier. As we are generating candidate entity mentions by using top $2$ most probable BIO sequences, there may be some overlapping entity mentions. For each pair of such overlapping candidate entity mentions (say $i$ and $j$), following rules are added so that at most one of them is a valid entity mention.

\vspace{-5mm}
{\footnotesize
\begin{eqnarray*}
!ETFinal(i,NONE) \Rightarrow ETFinal(j,NONE).\\
!ETFinal(j,NONE) \Rightarrow ETFinal(i,NONE).
\end{eqnarray*}
}
We assume candidate mentions generated from the second BIO sequence to be valid, only if they are involved in some valid relation other than $NULL$. Also, an invalid entity mention should not be involved is any non $NULL$ relation with any other mention. To ensure this desired consistency, following rules are added for each pair of candidate mentions ($i,j$) where one of them (say $i$) is generated from second BIO sequence.

\vspace{-5mm}
{\footnotesize
\begin{eqnarray*}
!RTFinal(i,j,NULL) \Rightarrow !ETFinal(i,NONE).\\
ETFinal(i,NONE) \Rightarrow RTFinal(i,j,NULL).\\
\end{eqnarray*}
}
After the inference, if the probability of $ETFinal(i,NONE)$ is the highest for any candidate mention $i$, then it is identified as an invalid mention. And because of above rules ensuring consistency, such mentions are never involved in any non $NULL$ relation.

\subsection{Additional Semantic Rules}
We explored the possibility of incorporating some domain knowledge by exploiting the easy and effective representability of the first-order logic. In order to incorporate the additional rules, we define following new predicates:\\
1. $CONS(entity,entity):$ 
$CONS(i,j)$ is true only when there is no other entity mention occurring in between the mentions $i$ and $j$ in a sentence.\\
2. $CONJ(entity,entity):$ 
$CONJ(i,j)$ is true only when there is a conjunction (i.e. connected through the dependency relations ``conj:and'' or ``conj:or'' in the dependency tree) between the two mentions $i$ and $j$.

\noindent\textbf{Using the knowledge of conjunctions:} When two entity mentions are connected through a conjunction (like \texttt{and}, \texttt{or}) and one of them is connected to a third entity mention with PHYS (i.e. located at) relation, then the other entity mention is also very likely to be connected to the third mention with PHYS relation. E.g. in the sentence fragment \texttt{troops in Israel and Syria}, a PHYS relation between \texttt{troops} and \texttt{Israel} implies another PHYS relation between \texttt{troops} and \texttt{Syria}. To incorporate this knowledge, following generic rules are added in MLNs of all sentences.

\vspace{-5mm}
{\footnotesize
\begin{eqnarray*}
RTFinal(x,y,PHYS) \wedge ((CONJ(y,z) \wedge CONS(y,z)) \vee (CONJ(z,y) \wedge CONS(z,y)))\\ \wedge 
ET(y,t) \wedge ET(z,t) \Rightarrow RTFinal(x,z,PHYS).
\end{eqnarray*}
\begin{eqnarray*}
RTFinal(x,y,PHYS) \wedge ((CONJ(w,x) \wedge CONS(w,x)) \vee (CONJ(x,w) \wedge CONS(x,w)))\\
\wedge ET(w,t) \wedge ET(x,t) \Rightarrow RTFinal(w,y,PHYS).
\end{eqnarray*}
}

\noindent\textbf{Linking entity mentions with same types:} The entity mentions linked through certain dependency relations tend to share the same entity type. E.g. in the sentence fragment \texttt{companies such as Nielsen}, the mentions \texttt{companies} and \texttt{Nielsen} are very likely to have the same entity type. This is one of the Hearst patterns~\cite{hearst1992automatic} to automatically identify hyponyms from text. If entity mentions $i$ and $j$ follow such a pattern, we add following rule to their sentence's MLN : $ETFinal(i,x) \Leftrightarrow ETFinal(j,x).$

\noindent\textbf{Using knowledge about relation types:} If an entity mention of type $PER$ is involved in a $EMPORG$ relation, then it is highly unlikely that the same person will be connected to any other mention with the $EMPORG$ relation. This is because any person can have at most one employer mentioned in a single sentence. To impose this constraint, we add following rule.

\vspace{-5mm}
{\footnotesize
\begin{eqnarray*}
RTFinal(x,y,EMPORG) \wedge (y \neq z) \wedge !RTFinal(y,z,IDN) \wedge !RTFinal(z,y,IDN)\\ \wedge ETFinal(x,PER) \Rightarrow !RTFinal(x,z,EMPORG) \wedge !RTFinal(z,x,EMPORG).
\end{eqnarray*}
}
\subsection{Joint Inference}
As described above, an MLN is created for a sentence using some {\em Generic Rules} with infinite weights and some sentence-specific rules. Given such an MLN, we are interested to know the most probable groundings of the {\em query} predicates given some specific groundings of {\em evidence} predicates. 
In our case, $ETFinal$ and $RTFinal$ are the {\em query} predicates and $ET$, $RTP$, $RTL$, $CONS$ and $CONJ$ are the {\em evidence} predicates. Inference over this MLN gives the probability of each possible grounding of the {\em query} predicates, conditioned on the given values of the {\em evidence} predicates. We used the default inference algorithm in Alchemy named ``Lifted Belief Propagation''~\cite{singla2008lifted}. For each candidate entity mention $i$, grounding of the predicate $ETFinal(i,E)$ with the highest probability is chosen and corresponding value of $E$ is its final entity type except the case when $E=NONE$. In that case, we do not identify the corresponding candidate mentions as a valid entity mention. Similarly, for each entity mention pair $(i,j)$, grounding of the predicate $RTFinal(i,j,R)$ with the highest probability  is chosen and corresponding $R$ value is its final relation type.

\section{Example}
In this section, we describe an example sentence where the joint inference helps in correcting the prediction errors by the individual classifiers. Consider the sentence from the ACE 2004 dataset: \texttt{she is the new chair of the black caucus.} 
In order to identify the candidate entity mentions, top 2 label sequences predicted by the CRF model are considered. 
\begin{enumerate}
\item {\tt she}/B {\tt is}/O {\tt the}/O {\tt new}/O {\tt chair}/O {\tt of}/O {\tt the}/O {\tt black}/O {\tt caucus}/B {\tt .}/O
\item {\tt she}/B {\tt is}/O {\tt the}/O {\tt new}/O {\tt chair}/B {\tt of}/O {\tt the}/O {\tt black}/O {\tt caucus}/B {\tt .}/O
\end{enumerate}
Table~\ref{tab:ex2} shows all the candidate entity mentions identified along with their IDs and predictions of the local entity classifier.
\vspace*{-5mm}
\begin{table}
\caption{Candidate entity mentions identified in the example sentence}
\vspace*{-5mm}
\label{tab:ex2}
{\footnotesize
\begin{center}
\begin{tabular}{|c|c|c|c|c|}
\hline
\textbf{ID} & \textbf{Entity Mention} & \textbf{From First BIO Sequence?} & \textbf{Predicted Type} & \textbf{Actual Type}\\
\hline
1 & \texttt{she} & Yes & PER & PER\\
2 & \texttt{chair} & No & PER & PER\\
3 & \texttt{caucus} & Yes & PER & ORG \\
\hline
\end{tabular}
\end{center}
}
\end{table}
It can be observed that mention ID 2 is generated from the second best BIO sequence and hence will be considered a valid mention only if it is involved in a relation with some other mention. Moreover, the entity type predicted for the mention ID 3 (\texttt{caucus}) is incorrect. This error propagates to the relation classification with ``pipeline'' classifier predicting relation between \texttt{chair} and \texttt{caucus} to be IDN instead of EMP-ORG. But the local classifier predicts the correct relation type EMP-ORG for this pair as it is not using the entity type features. The first-order logic rules for this sentence's MLN are shown in the Table~\ref{tab:ex_rules}. The LOR (log of odds ratio) weights assignment strategy is used. In case of soft constraints, the number preceding each rule indicates its weight. No weight is explicitly specified for the hard constraints and they always end with a period.
\vspace*{-5mm}
\begin{table}
\caption{First-order logic rules for the MLN of example sentence}
\vspace*{-5mm}
\label{tab:ex_rules}
{\footnotesize
\begin{center}
\begin{tabular}{|p{5.9cm}|p{6.7cm}|}
\hline
\textbf{Rules induced by the local entity classifier} & \textbf{Rules for identifying valid/invalid entity mentions}\\
\hline
$6.13\hspace{2mm}ET(1$,$PER)\Leftrightarrow ETFinal(1$,$PER)$ & $6.13\hspace{2mm}!ETFinal(1$,$NONE)$\\
\cline{1-1}
$-0.93\hspace{2mm}ET(2$,$PER)\Leftrightarrow ETFinal(2$,$LOC)$ & $0.71\hspace{2mm}ETFinal(2$,$NONE)$\\
$-0.89\hspace{2mm}ET(2$,$PER)\Leftrightarrow ETFinal(2$,$ORG)$ & $0.15\hspace{2mm}!ETFinal(3$,$NONE)$\\
$-0.71\hspace{2mm}ET(2$,$PER)\Leftrightarrow ETFinal(2$,$PER)$ & $ETFinal(2$,$NONE)\Rightarrow RTFinal(1$,$2$,$NULL).$\\
\cline{1-1}
$-0.53\hspace{2mm}ET(3$,$PER)\Leftrightarrow ETFinal(3$,$ORG)$ & $!RTFinal(1$,$2$,$NULL)\Rightarrow !ETFinal(2$,$NONE).$\\
$0.15\hspace{2mm}ET(3$,$PER)\Leftrightarrow ETFinal(3$,$PER)$ & $ETFinal(2$,$NONE)\Rightarrow RTFinal(2$,$3$,$NULL).$\\
 & $!RTFinal(2$,$3$,$NULL) \Rightarrow!ETFinal(2$,$NONE).$\\
\hline
\hline
\multicolumn{2}{|l|}{\bf Rules induced by the local and pipeline relation classifiers}\\
\hline
\multicolumn{2}{|l|}{$3.37\hspace{2mm}RTL(1,2,IDN) \Leftrightarrow RTFinal(1,2,IDN)$}\\
\multicolumn{2}{|l|}{$2.99\hspace{2mm}RTP(1,2,IDN) \Leftrightarrow RTFinal(1,2,IDN)$}\\
\hline
\multicolumn{2}{|l|}{$1.52\hspace{2mm}RTL(1,3,NULL) \Leftrightarrow RTFinal(1,3,NULL)$}\\
\multicolumn{2}{|l|}{$-1.66\hspace{2mm}RTL(1,3,NULL) \Leftrightarrow RTFinal(1,3,IDN)$}\\
\multicolumn{2}{|l|}{$0.35\hspace{2mm}RTP(1,3,NULL) \Leftrightarrow RTFinal(1,3,NULL)$}\\
\multicolumn{2}{|l|}{$-1.63\hspace{2mm}RTP(1,3,NULL) \Leftrightarrow RTFinal(1,3,IDN)$}\\
\hline
\multicolumn{2}{|l|}{$-1.80\hspace{2mm}RTL(2,3,EMPORG) \Leftrightarrow RTFinal(2,3,PHYS)$}\\
\multicolumn{2}{|l|}{$-1.09\hspace{2mm}RTL(2,3,EMPORG) \Leftrightarrow RTFinal(2,3,IDN)$}\\
\multicolumn{2}{|l|}{$0.24\hspace{2mm}RTL(2,3,EMPORG) \Leftrightarrow RTFinal(2,3,EMPORG)$}\\
\multicolumn{2}{|l|}{$-0.46\hspace{2mm}RTP(2,3,IDN) \Leftrightarrow RTFinal(2,3,IDN)$}\\
\hline
\end{tabular}
\end{center}
}
\end{table}
\vspace*{-10mm}
\begin{table}\center
\caption{MLN inference output for entity types}
\label{tab:etf}
{\footnotesize
\begin{tabular}{|c|c|c|}
\hline
{\tt she} (ID 1) & {\tt chair} (ID 2) & {\tt caucus} (ID 3) \\
\hline
$ETFinal(1,PER)=${\bf0.99} & $ETFinal(2,PER)=${\bf0.92} & $ETFinal(3,PER)=$0.35\\
$ETFinal(1,GPE)=$0.01 & $ETFinal(3,GPE)=$0.03 & $ETFinal(3,ORG)=${\bf0.39}\\
 & $ETFinal(2,GPE)=$0.02 & $ETFinal(3,NONE)=$0.14\\
 & $ETFinal(2,NONE)=$0.01 & $ETFinal(3,FAC)=$0.03\\
\hline
\end{tabular}
}
\end{table}
\begin{table}\center
\caption{MLN inference output for relation types}
\label{tab:rtf}
{\footnotesize
\begin{tabular}{|c|c|c|}
\hline
{\tt (she,chair)} & {\tt (she, caucus)} \\
\hline
$RTFinal(1,2,IDN)=${\bf0.92} & $RTFinal(1,3,EMPORG)=$0.02\\
$RTFinal(1,2,PHYS)=$0.01 & $RTFinal(1,3,PERSOC)=$0.02\\
$RTFinal(1,2,ART)=$0.01 & $RTFinal(1,3,OTHERAFF)=$0.02\\
$RTFinal(1,2,OTHERAFF)=$0.02 & $RTFinal(1,3,NULL)=${\bf0.90}\\
$RTFinal(1,2,NULL)=$0.01 & $RTFinal(1,3,IDN)=$0.02\\
\hline
{\tt (chair,caucus)} & \\
\hline
$RTFinal(2,3,EMPORG)=${\bf0.33}& \\
$RTFinal(2,3,PHYS)=$0.10 & \\
$RTFinal(2,3,GPEAFF)=$0.10 & \\
$RTFinal(2,3,OTHERAFF)=$0.09 & \\
$RTFinal(2,3,IDN)=$0.20 & \\
\hline
\end{tabular}
}
\end{table}

The joint inference combines the evidence from the above three classifiers and generates a globally consistent output. The outputs for the query predicates $ETFinal$ and $RTFinal$ are shown in the Tables~\ref{tab:etf} and~\ref{tab:rtf}, respectively. The predicate groundings which have negligible probability are not shown. Here, it can be observed that the entity mention \texttt{chair} (ID 2) has been correctly identified as a valid mention and the type of entity mention \texttt{caucus} has been correctly predicted as ORG. Also the correct relation type of EMP-ORG between \texttt{chair} and \texttt{caucus} has been chosen as the global prediction.

\section{Experimental Analysis}
In order to demonstrate the effectiveness of our approach, we compare its performance with other approaches which have reported their results for end-to-end relation extraction on ACE 2004 dataset~\footnote{We have not yet acquired a more recent ACE 2005 dataset}.
For fair comparison, we follow the same assumptions made by Chan and Roth~\cite{chan2011exploiting} and Li and Ji~\cite{li2014incremental}, i.e. ignoring the DISC relation, not treating implicit relations as false positives and using coarse entity and relation types.
All the results are obtained by 5-fold cross-validation on ACE-2004 data. Note that the actual folds used by each algorithm may differ.

\begin{table}[!h]\small
\caption{Results on the ACE 2004 dataset (Micro-averaged, 5-fold cross-validation)}
\label{tab:results}
\begin{tabular}{|l|ccc|ccc|ccc|}
\hline
{\bf Approach} & \multicolumn{3}{|c|}{\bf Entity Extraction} & \multicolumn{3}{|c|}{\bf Relation Extraction} & \multicolumn{3}{|c|}{\bf Entity+Relation}\\
  & \hspace{4mm}{\bf P}\hspace{4mm} & {\bf R} & {\bf F} & \hspace{4mm}{\bf P}\hspace{4mm} & {\bf R} & {\bf F} & \hspace{4mm}{\bf P}\hspace{4mm} & {\bf R} & {\bf F}\\
\hline
Local Classifiers & 80.9 & 77.6 & 79.2 & 53.2 & 43.9 & 48.1 & 46.2 & 38.1 & 41.8 \\
Pipeline Classifier &  &  &  & 53.3 & 46.4 & 49.6 & 48.7 & 42.5 & 45.4 \\
Chan and Roth~\cite{chan2011exploiting} &  &  &  &  42.9 & 38.9 & 40.8 &  &  &  \\
Li and Ji~\cite{li2014incremental} & 83.5 & 76.2 & 79.7 &  64.7 & 38.5 & 48.3 & 60.8 & 36.1 & 45.3 \\
Miwa and Bansal~\cite{miwa2016end} & 83.3 & 79.2 & {\bf81.2} &  &  &  & 56.1 & 40.8 & {\bf 47.2} \\
MLN (LOR) & 79.3 & 79.9 & 79.6 & 56.2 & 45.2 & 50.1 & 50.6 & 40.8 & 45.2 \\
MLN (LOR)+Rules & 79.3 & 80.0 & 79.6 & 56.6 & 45.1 & 50.2 & 51.0 & 40.6 & 45.2 \\
MLN (CM) & 78.9 & 80.1 & 79.5 & 57.2 & 45.2 & 50.5 & 51.6 & 40.8 & 45.6 \\
MLN (CM)+Rules & 79.0 & 80.1 & 79.5 & 57.9 & 45.6 & {\bf51.0} & 52.4 & 41.3 & 46.2 \\
\hline

\end{tabular}
\end{table}

Comparative performances of all the approaches are shown in the table~\ref{tab:results}. A true positive for the task of entity extraction means that an entity mention has been correctly identified as the valid mention and also its type has been identified correctly. A true positive for the task of relation extraction means that for a pair of valid entity mentions, its relation type (except for special relation types $NULL$ and $IDN$) has been identified correctly. For entity+relation extraction, a stricter criteria is used where a true positive means that for a pair of valid entity mentions, not only its relation type is identified correctly but types of both the mentions are also identified correctly. Even if any one of these predictions is incorrect, we consider it as a false positive for the predicted combination of entity types and relation type and also as a false negative for the true combination of entity types and relation type.

It can be observed that MLN inference with CM (Constant Multiplier) weights assignment strategy performs better that the LOR (Log of Odds Ratio) in case of relation extraction whereas for entity extraction LOR strategy is better. Addition of semantic rules (discussed in the Section 3.5) results in better performance for both the strategies. Also, we can observe that MLN (CM) with semantic rules comfortably outperforms the individual classifiers: local entity classifier, local relation classifier and ``pipeline'' relation classifier. 
In case of end-to-end relation extraction, our approach outperforms the approaches of Chan and Roth~\cite{chan2011exploiting} and Li and Ji~\cite{li2014incremental} on the ACE 2004 dataset and also achieves a comparable performance as compared to Miwa and Bansal~\cite{miwa2016end}. We also achieve comparable performance in case of entity extraction as compared to Li and Ji~\cite{li2014incremental} but underperform in comparison with Miwa and Bansal~\cite{miwa2016end}.

\section{Related Work}
Previous work on joint extraction of entities and relations can be broadly classified into 5 categories : i) Integer Linear Programming (ILP) based approaches~\cite{roth2004linear,roth2007global}, ii) Probabilistic Graphical Models~\cite{roth2002probabilistic,singh2013joint}, iii) Card-pyramid parsing~\cite{kate2010joint}, iv) Structured Prediction~\cite{li2014incremental,li2014constructing,miwa2014modeling} and v) Recurrent Neural Network (RNN) based model~\cite{miwa2016end}. Our approach is similar to ILP based approaches, but we use MLNs for joint inference which provide much better representation to incorporate complex domain knowledge as compared to ILP. For example, the rules defined in the section 3.5 are quite easy to incorporate using first-order logic but the same would be cumbersome in ILP. The approaches by Singh et al.~\cite{singh2013joint} and Li and Ji~\cite{li2014incremental} not only carry out joint ``inference'' but also create a joint ``model'' where the parameters for both the tasks are learned jointly. 

Zhang et al.~\cite{zhang2012ontological} used Markov Logic rules to perform {\em Ontological Smoothing}. The concept of {\em Ontological Smoothing} is to find a mapping from a user-specified target relation to a background knowledge base. This mapping is then used to generate extra training data for distant supervision. Similar to our approach, they also use Markov logic rules to ensure consistency between relation types and entity types. One major difference is that the relation types used by them were quite specific and not as general as ACE 2004 relation types. Zhu et al.~\cite{zhu2009statsnowball} also used MLNs but they addressed a relation extraction problem which is bit different from the ACE 2004 RDC task. It requires the explicit mention of relation in the form of words other than the words inside entity mentions. This is not always true for ACE 2004 relations. For example, EMP-ORG relation holds between \texttt{Indian} and \texttt{soldiers} in the sentence \texttt{\underline{Indian} \underline{soldiers} attacked the terrorists.}

\section{Conclusion and Future Work}
We described the problem of end-to-end relation extraction and the need to jointly address its sub-tasks of entity and relation extraction. 
We proposed a new approach for joint extraction of entity mentions and relations at the sentence level, which uses joint inference in Markov Logic Networks (MLN). We described in detail about the domains, predicates and first-order logic rules used to create an MLN for a sentence. We also explored how the effective representability of first-order logic can be used to incorporate various semantic rules and domain knowledge. 
Finally, we demonstrated better than the state-of-the-art end-to-end relation extraction performance on the standard dataset of ACE 2004.

In future, we plan to analyze the two weights assignment strategies (CM and LOR) in detail and develop deeper understanding of pros and cons of each one. Also, we have tried only a small number of additional semantic rules. In future, we wish to take advantage of the first-order logic framework to incorporate deeper semantic knowledge. Another important direction to explore is about learning the weights of first-order logic rules automatically.
\bibliographystyle{splncs03}
\bibliography{ref}
\end{document}